\documentclass[letterpaper]{article} 
\usepackage[submission]{aaai24}  
\usepackage{float}
\usepackage{times}  
\usepackage{helvet}  
\usepackage{courier}  
\usepackage[hyphens]{url}  
\usepackage{graphicx} 
\usepackage{natbib}  
\usepackage{caption} 

\usepackage{amsmath}
\usepackage{amssymb}
\usepackage{graphicx}
\usepackage{xcolor}
\usepackage{booktabs}
\usepackage{makecell}
\usepackage{xspace}
\usepackage{multirow}
\usepackage{threeparttable}
\usepackage{algorithmic}
\usepackage{algorithm}
\usepackage{cleveref}
\usepackage{bbding}
\usepackage{makecell}

\newcommand{\eg}{\emph{e.g.},\xspace}
\newcommand{\ie}{\emph{i.e.},\xspace}
\newcommand{\etc}{etc.\xspace}

\newcommand\figref[1]{Figure~\ref{#1}}

\newcommand\tabref[1]{Table~\ref{#1}}

\newcommand{\fakeparagraph}[1]{\vspace{1mm}\noindent\textbf{#1.}}

\newcommand{\sysnamefull}{\textbf{D}ialogue-comprised \textbf{P}olicy-gradient-based \textbf{D}iscrete \textbf{P}rompt \textbf{O}ptimization}
\newcommand{\sysname}{$\textsc{DP}_2\textsc{O}$}
\newcommand{\metricname}{$\mathrm{SUE}$}

\usepackage{newfloat}
\usepackage{listings}
\DeclareCaptionStyle{ruled}{labelfont=normalfont,labelsep=colon,strut=off} 
\floatstyle{ruled}
\newfloat{listing}{tb}{lst}{}
\floatname{listing}{Listing}
\title{\textbf{Dialogue for Prompting: a Policy-Gradient-Based Discrete Prompt Generation for Few-Shot Learning}}
\author{
    Chengzhengxu Li, Xiaoming Liu\textsuperscript{\rm $\star$}, Yichen Wang, Duyi Li, Yu Lan, Chao Shen
}
\affiliations{

     Faculty of Electronic and Information Engineering, Xi’an Jiaotong University \\
     \textrm{\{czx.li, yichen.wang, liduyi\}@stu.xjtu.edu.cn, \{xm.liu, ylan2020, chaoshen\}@xjtu.edu.cn}
    
    \textrm{\textsuperscript{\rm $\star$}Corresponding author}
    
}

\usepackage{bibentry}

\begin{document}

\maketitle

\begin{abstract}
Prompt-based pre-trained language models (PLMs) paradigm has succeeded substantially in few-shot natural language processing (NLP) tasks.
However, prior discrete prompt optimization methods require expert knowledge to design the base prompt set and identify high-quality prompts, which is costly, inefficient, and subjective.
Meanwhile, existing continuous prompt optimization methods improve the performance by learning the ideal prompts through the gradient information of PLMs, whose high computational cost, and low readability and generalizability are often concerning.
To address the research gap, we propose a \sysnamefull{} (\sysname{}) method. 
We first design a multi-round dialogue alignment strategy for readability prompt set generation based on GPT-4.
Furthermore, we propose an efficient prompt screening metric to identify high-quality prompts with linear complexity. 
Finally, we construct a reinforcement learning (RL) framework based on policy gradients to match the prompts to inputs optimally. 
By training a policy network with only \textbf{0.62M} parameters on the tasks in the few-shot setting, \sysname{} outperforms the state-of-the-art (SOTA) method by \textbf{1.52\%} in accuracy on average on four open-source datasets.
Moreover, subsequent experiments also demonstrate that \sysname{} has good universality, robustness and generalization ability.\footnote{Code and data are avaliable at \url{https://github.com/czx-li/DP2O}.}
\end{abstract}

\begin{figure}[h]
\centering
\includegraphics[width=1.03\columnwidth]{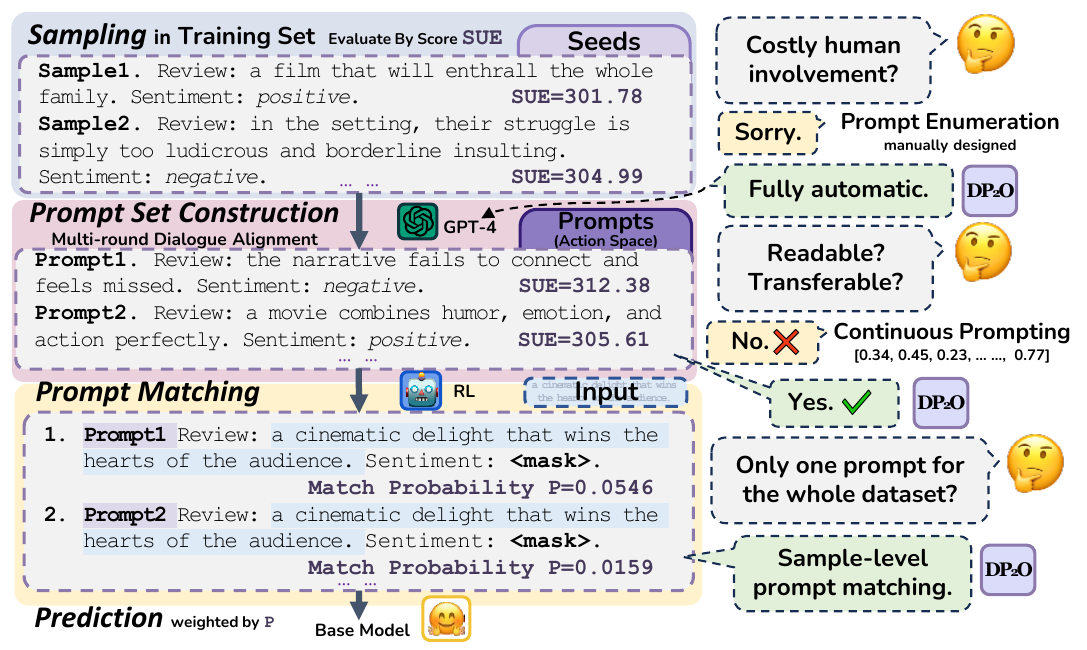} 
\caption{An illustration of the procedure and innovation Q\&A of \sysname{}. The procedure includes 
1) Sampling the seed set from the training set via \metricname{}; 
2) Constructing the prompt by multi-round dialogue with GPT-4 to align the inputs with the whole training set's distribution;
3) Employing an RL agent to match prompts with inputs to predict probabilistically;
4) Feeding all prompt-input pairs to a base PLM model for downstream tasks and ensemble predictions by probability weighting.}
\label{introfig}
\end{figure}

\section{Introduction}
With the continuous development of pre-trained language models (PLMs) \cite{liu2019roberta, touvron2023llama, anil2023palm}, 
\eg ChatGPT \cite{OpenAI2022ChatGPT} and GPT-4 \cite{OpenAI2023GPT4TR}, prompt-based methods have shown significant rising competitiveness in few-shot downstream tasks \cite{schick2020exploiting, schick2020s}. 
Unlike the traditional fine-tuning methods, which require the design of additional neural network heads according to downstream tasks, the prompt-based methods join particular extra texts to inputs to transfer downstream tasks into mask-filling tasks. 
The prompt matches the downstream task with the model's pre-training task, and the potential of the PLMs can be more comprehensively scheduled.
However, PLMs are extremely sensitive to prompts \cite{holtzman2019curious, lester2021power}. 
Minor gaps with the same semantics in prompts may lead PLMs to completely different performances. 
Therefore, one core issue of the prompt-based methods is finding high-quality prompts to promote the performance of PLMs.

Currently, prompt optimization methods can be divided into two categories: \textit{discrete prompt optimization} and \textit{continuous prompt optimization}. 

Due to the discrete nature of the text, prompts can not be directly optimized by using the gradient information from PLMs. 
Therefore, previous \textit{discrete prompt optimization} methods heavily relied on the manually designed basic prompt sets and prompt templates \cite{jiang2020can, yuan2021bartscore, haviv2021bertese, davison2019commonsense}. 
Moreover, lacking clear evaluation metrics, prior works often use the supervision gain of training-set prompt as a screening metric during optimization \cite{zhou2022large, gao2020making}. 
Therefore, \textit{discrete prompt optimization} methods usually necessitate a number of labeled data, which contradicts the few- or zero-shot learning objectives, and overlooks the potential impact of prompts on the output distribution, and the further effect on the performance of PLMs. 

\begin{figure*}[t]
\centering
\includegraphics[width=\linewidth]{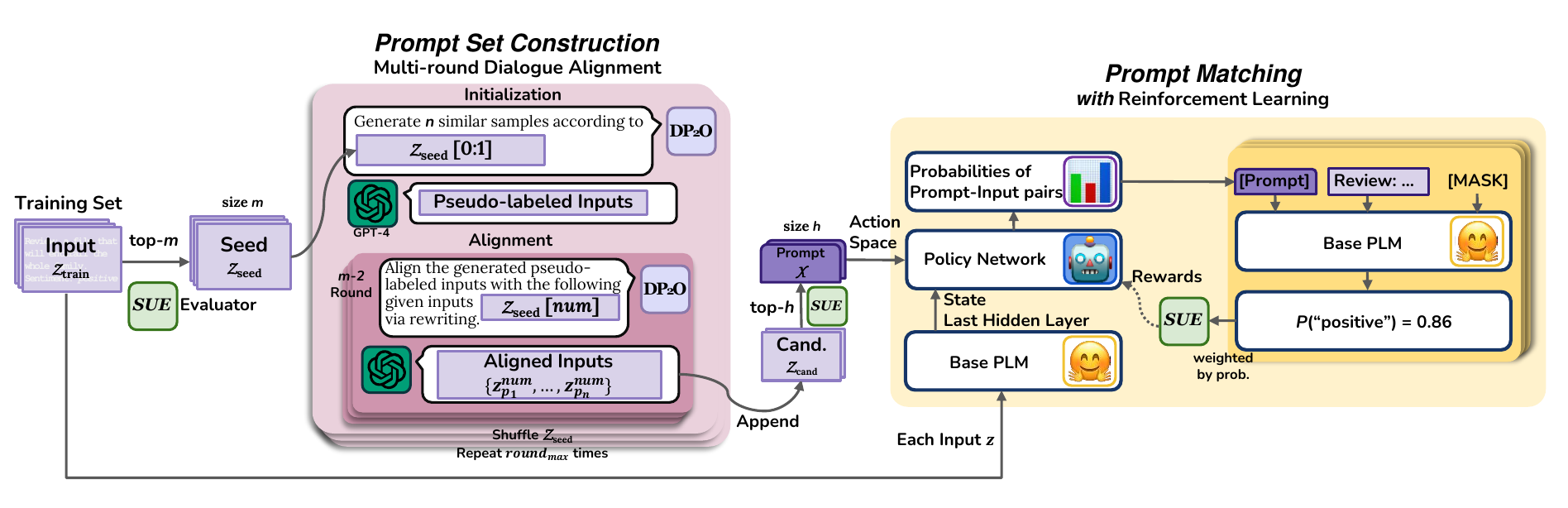} 
\vspace{-10pt}
\caption{{Overview of \sysname{}.
In the \textit{prompt set construction} stage, we use the multi-round dialogue alignment strategy to generate high-quality discrete prompts continuously. Given the seed inputs $\mathcal{Z}_{\textrm{seed}}$ with top-$m$ \metricname{} score, \sysname{} have a conversation with GPT-4, which has $round_{max}$ times outer loop and $m - 2$ times inner loop, to align inputs semantics with the training set. Then \sysname{} apply the assessment metric \metricname{} to sort the prompts after dialogue to obtain the final candidate set $\mathcal{Z}_{\mathrm{cand}}$. 
We filter top-$h$ candidates as the final prompt set based on \metricname{} score.
In the \textit{prompt matching} stage, we build a reinforcement learning framework to match the appropriate prompt from $\mathcal{X}$ for each input $z$ from $\mathcal{Z}_{\textrm{train}}$ with probability. The prompt-input pairs are fed into the base PLM to predict downstream tasks. The final prediction is the probability-weighted output of all pairs.}}
\label{overall}
\end{figure*}

Meanwhile, the \textit{continuous prompt optimization} methods abandon the text structure of prompts and improve the performance of PLMs by directly optimizing token embedding at specific locations \cite{vu2021spot, li2021prefix, an2022input, qian2022controllable}. 
Although these methods can directly use gradient information to guide the optimization direction of continuous prompts, the computational cost is often exceedingly expensive. 
Besides, continuous prompts usually lack readability and are hardly used across different PLMs.

Towards these challenges, we propose a \sysnamefull{} method, named  \sysname{}. 
As shown in \figref{introfig}, \sysname{} mainly consists of two stages: \textit{prompt set construction} and \textit{prompt matching}. 
In the \textit{prompt set construction} stage, we propose a prompt set generation method with a multi-round dialogue alignment strategy by employing the dialogue characteristics of GPT-4, one of the current most capable PLMs on dialogue.
Meanwhile, we introduce an innovative prompt quality assessment metric, \ie Supervised \& Unsupervised Entropy Metric (\metricname{}), which comprehensively considers the supervised and unsupervised impact of prompts on PLMs with linear complexity and facilitates output distribution balance and accuracy in downstream tasks. 
In the \textit{prompt matching} stage, 
we propose a reinforcement learning (RL) framework, which employs a policy network to select appropriate input prompts. 
The prompts, without breaking textual semantics, ensure their readability and transferability across different PLMs.
Finally, the downstream task is completed through ensemble decision-making.
Extensive experiments show \sysname{} is significantly superior to baseline and SOTA methods,
\eg \sysname{} achieve an average improvement of \textbf{1.52\%} in accuracy across four public datasets with only a \textbf{10.86\%} training time of the SOTA method RLPrompt \cite{deng2022rlprompt}.
Furthermore, we implement ablation and analysis experiments to demonstrate the effectiveness, robustness and generalization of \sysname{}. 

In summary, our contributions are summarized as follows:
\begin{itemize}
\item \textbf{Novel Generation Strategy:} We generate the prompt set via the multi-round dialogue alignment strategy, aiming at reducing the cost of human involvement in prompting.
\item \textbf{Linear Evaluation Metric:} We additionally consider unsupervised information of PLMs prediction in prompts evaluation, proposing a new metric to screen out excellent prompts with linear complexity.
\item \textbf{Precise Prompt Matching:} We apply RL techniques to achieve sample-level discrete prompt optimization, further improving the performance of PLMs on downstream tasks.
\item \textbf{Outstanding Task Performance:} Experiments on four public datasets show that \sysname{} effectively improves the performance of PLMs under few-shot settings with readability, robustness, generalization, and universality. 
\end{itemize}

\section{Methodology}
The main workflow of \sysname{} can be mainly divided into two stages: \textit{prompt set construction} and \textit{prompt matching}, as shown in \figref{overall}.

\subsection{Prompt Set Construction Stage}
\fakeparagraph{Evaluation Metric}
Most prevailing methods utilize the aggregate accuracy of the prompt on the dataset as their sole metric for assessment, which neglects the impact of the distribution of labels in the dataset.
\citet{lu2021fantastically} find that significantly imbalanced prediction distributions typically characterize underperforming prompts. 
To this end, we introduce a novel evaluation metric termed \textbf{S}upervised \& \textbf{U}nsupervised \textbf{E}ntropy metric (\metricname{}).
\metricname{} aims to provide a more comprehensive appraisal for prompts by additionally considering global balance beyond local accuracy.

\metricname{} consists of two parts: \textit{supervision score} $S_{\textrm{sup}}$ and \textit{unsupervised score} $S_{\textrm{uns}}$. 
Given a prompt $x$ and input set ${\mathcal{Z}}$, for each input $z_i \in \mathcal{Z}$, we first calculate the difference of the probability $p_{\textrm{LM}}$ that the $z_i$ is correctly labeled $c_{i}$ and wrongly labeled as $c_{\textrm{else}}$ by a base PLM. 
Here $c_{\textrm{else}} \in \mathcal{C}\setminus \{c_i\}$ exactly, and $\mathcal{C}$ is the label space of the input.
Then the supervision score $S_{\textrm{sup}}$ of prompt $x$ is defined as:
\begin{align}
S_{\textrm{sup}}(x,\mathcal{Z}) =\sum_{z_{i}\in{\mathcal{Z}}} (p_{\textrm{LM}}(c_{i}|x,z_{i})-p_{\textrm{LM}}(c_{\textrm{else}}|x,z_{i}))
\end{align}

To prevent some prompts from causing PLMs to be overly biased on all inputs, \metricname{} selects the prompts which guide PLMs to output a more balanced pseudo-label distribution across all given inputs. 
Given a prompt $x$,  we calculate a entropy value $\mathbb{H}(.)$ of each input $z_i$, then add $\mathbb{H}(.)$ of each input as $S_{\textrm{uns}}$ for the whole input set $\mathcal{Z}$:
\begin{gather}
    \mathbb{H}(x,z_{i})=\sum\nolimits_{c_{i}\in{\mathcal{C}}}-p_{\textrm{LM}}(c_{i}|x,z_{i})\log p_{\textrm{LM}}(c_{i}|x,z_{i})  \\
    S_{\textrm{uns}}(x,\mathcal{Z})= \sum\nolimits_{z_{i}\in{\mathcal{Z}}}\mathbb{H}(x,z_{i}) 
\end{gather}

Finally, we have our evaluation metric \metricname{} to assess the quality of prompts as
\begin{align}
\mathrm{SUE}(x,\mathcal{Z})=\lambda_{1}S_{\textrm{sup}}(x,\mathcal{Z})+\lambda_{2}S_{\textrm{uns}}(x,\mathcal{Z})
\end{align}
where $\lambda_{1}$ and $\lambda_{2}$ are the weights to balance the supervised score $S_{\textrm{sup}}$ and the unsupervised score $S_{\textrm{uns}}$. 
For the input set $\mathcal{Z}$ encompassing multiple inputs, the metric \metricname{} characterizes the holistic quality of the prompt. 
Higher \metricname{} represents the better capability on the specific downstream task (derived from $S_{\textrm{sup}}$) and more benign confidence on all inputs (derived from $S_{\textrm{uns}}$). 
Meanwhile, when $\mathcal{Z}$ only comprises a single input, \metricname{} can elucidate the degree of match between the prompt and the input.

\subsubsection{Prompt Set Generation.}
Existing discrete prompt optimization methods, such as Black-Box Tuning \cite{sun2022black} and GrIPS \cite{prasad2022grips}, mostly require text editing based on manually designed prompts and vocabularies. Different from these methods, \sysname{} leverages GPT-4 as a dialogue model to generate pseudo-label inputs which approximate the dataset distribution, utilizing only a limited set of training data. 
Then the inputs are used as prompt examples for downstream tasks. 
Notably, \citet{min2022rethinking} indicate that the label authenticity of these pseudo-label prompt examples has little impact on the performance of PLMs. 
Hence, for \sysname{}, we do not validate the authenticity of the labels of the inputs generated by GPT-4, which further eliminates the necessity for human annotation. 
Our experiments also show that, without verifying the authenticity of labels, \sysname{} can still achieve praiseworthy and competitive performance to other methods.


\begin{algorithm}
\small
	\caption{Prompt Set Construction of \sysname{}} 
	\label{alg:1}
	\begin{algorithmic}[1]
		\REQUIRE Few-shot training set $\mathcal{Z}_{\textrm{train}}$ including inputs and labels, label space $\mathcal{C}$, base PLM and access to GPT-4 API. 
            \STATE $\mathcal{Z}_{\textrm{seed}} \leftarrow $ top-$m$ inputs in $\mathcal{Z}_{\textrm{train}}$ via $\mathrm{SUE}(z_{i}, \mathcal{Z}_{\textrm{train}})$, $z_i \in \mathcal{Z_{\textrm{train}}}$.   
            \begin{center}
  \textit{**** outer-loop begins: multi-round dialogue ****}
            \end{center}
            \STATE $round \leftarrow 0 $
            \WHILE{$round< round_{max}$} 
                \STATE Random shuffle $\mathcal{Z}_{\textrm{seed}}$ .
                \STATE Input $\mathcal{Z}_{\textrm{seed}}[0:1]$ to GPT-4 with prefix introduction of task.
                \STATE GPT-4 output $n$ pseudo-labeled inputs $\{z_{p_1}^1, ..., z_{p_n}^1\}$.
                \begin{center}
              \textit{**** inner-loop begins: one dialogue round ****}
                 \end{center}
                \STATE Initialize number of used inputs in $\mathcal{Z}_{\textrm{seed}}$ $num \leftarrow 2$.
                \WHILE{$num < m$}
                    \STATE Input $\mathcal{Z}_{\textrm{seed}}[num]$ to GPT-4, asking it rewrite the previous $\{z_{p_1}^{num-1}, ..., z_{p_n}^{num-1}\}$.
                    \STATE GPT-4 output $\{z_{p_1}^{num}, ..., z_{p_n}^{num}\}$.
                    \STATE $num \leftarrow num + 1$
                \ENDWHILE
                \STATE Append $\{z_{p_1}^{{m-1}}, ..., z_{p_n}^{{m-1}}\}$ to $\mathcal{Z}_{\textrm{cand.}}$
            \ENDWHILE 
            \STATE $\mathcal{X} \leftarrow$ top-$h$ inputs of $\mathcal{Z}_{\textrm{cand.}}$ via $\mathrm{SUE}(z_{i}, \mathcal{Z}_{\textrm{train.}})$ , $z_i \in \mathcal{Z}_{\textrm{cand.}}$
        \ENSURE Readable and high-quality prompt set $\mathcal{X}$. 
	\end{algorithmic}
\end{algorithm}

With the prevalence of PLMs aiming at chatting, dialogue is an effective way to input multi-inputs to models. 
Instead of concatenating them into long sequence text, dialogue strategy can ease the forgetness of PLMs caused by the sliding window.
As shown in Algorithm \ref{alg:1}, we utilize dialogue to gradually align our prompts with the distribution of PLM to reduce the potential threat of biased prediction.

Given a training set denoted as $\mathcal{Z}_{\textrm{train}}$, we first individually evaluate each input $z_{i}\in \mathcal{Z}_{\textrm{train}}$ by score $\mathrm{SUE}(z_{i}, \mathcal{Z}_{\textrm{train}})$. 
\metricname{} signifies the input $z_i$'s efficacy within the given set $\mathcal{Z}_{\textrm{train}}$. 
We rank $\mathcal{Z}_{\textrm{train}}$ set in descending order based on the  $\mathrm{SUE}(z_{i}, \mathcal{Z}_{\textrm{train}})$. 
We then select the top-$m$ inputs to form the seed set $\mathcal{Z}_{\textrm{seed}} = \{z_{\textrm{seed}_1}, z_{\textrm{seed}_2}, ..., z_{\textrm{seed}_m}\}$.

Subsequently, utilizing GPT-4, we generate pseudo-label inputs that mirror the distribution of the prompts within the $\mathcal{Z}_{\textrm{seed}}$. 
Initially, GPT-4 randomly takes inputs from $\mathcal{Z}_{\textrm{seed}}$ and then begins a \textit{dialogue round} (outer-loop).

In one \textit{dialogue round}, we first generate $n$ pseudo-labeled inputs $\{z_{p_1}^1, ..., z_{p_n}^1\}$ based on any two inputs from $\mathcal{Z}_{\textrm{seed}}$. 
Then we randomly select one of the remaining inputs from $\mathcal{Z}_{\textrm{seed}}$ into dialogue to guide GPT-4 to polish the previously generated pseudo-labeled inputs $\{z_{p_1}^1, ..., z_{p_n}^1\}$ and get corresponding $\{z_{p_1}^2, ..., z_{p_n}^2\}$. 
Repeated the polishing stage (inner-loop) $m-2$ times until all $z_{\textrm{seed}}$ inputted to the dialogue once. Then we get $\{z_{p_1}^{m-1}, ..., z_{p_n}^{m-1}\}$ and append them in candidate set $\mathcal{Z}_{\textrm{cand}}$.

However, the order of conversation might impact dialogue alignment.
We suggest re-ordering the conversation multi-time to counteract the effect of the order. 
Thus, we shuffle $\{z_{\textrm{seed}_1}, z_{\textrm{seed}_2}, ..., z_{\textrm{seed}_m}\}$ and start a new \textit{dialogue round} (outer-loop), aiming at minimizing the impact of order in dialogue.
After finishing all $round_{max}$ times \textit{dialogue rounds} (outer-loop), we compute the score \metricname{} of all $n \times round_{max}$ inputs in $\mathcal{Z}_{\textrm{cand}}$. 
Then, we select the top-$h$ inputs from the candidate set $\mathcal{Z}_{\textrm{cand}}$ as the final prompt set $\mathcal{X}$. 
From now on, the selected inputs go into the role of prompt.

Overall, we introduce the \textit{multi-round dialogue alignment} strategy, optimizing GPT-4's utility in prompt generation and leveraging its inherent dialogue characteristic.

\subsection{Prompt Matching Stage}

Previous studies have underscored the high sensitivity of PLMs to prompts \cite{radford2018improving, dathathri2019plug, raffel2020exploring}.  
Traditional methods tend to rely on either random selection or a simple cosine similarity metric between the input and prompt embedding for selection \cite{gao2020making}. 
This leads to an under-exploration of the prompt space limiting the potential performance enhancements in complex tasks. 
While for the recently emerging RL-based methods, the complexity associated with brute-force searching escalates exponentially with data size growth. 
Hence, efficiently matching appropriate prompts for each input is a highly challenging task.


\begin{algorithm}
\small
	\caption{Prompt Matching of \sysname{}} 
	\label{alg:2}
	\begin{algorithmic}[1]
		\REQUIRE Training set $\mathcal{Z}_{\textrm{train}}$ of size $T$, testing set $\mathcal{Z}_{\textrm{test}}$, base PLM, the prompt set $\mathcal{X}$ constructed by Algorithm \ref{alg:1}.
            \begin{center}
  \textit{**** training the RL model ****}
            \end{center}
            \STATE Initialize the policy network $\pi_{\theta}$  parameters $\theta$ and $epoch \leftarrow 0$.
            \WHILE{$epoch<epoch_{max}$}    
                \FOR{step $t$ \textbf{in} $[1,..., T]$}
                    \STATE Get state $s_{t} \leftarrow \mathrm{PLM}(z_{t})$.
                    \STATE Run policy network $\pi_{\theta}(a_{t}|s_{t})$ to take an action $a_t$  to select a prompt $x_{t}$.
                    \STATE Calculate reward by SUE, \ie  $r_{t}\leftarrow\mathrm{SUE}(x_{t},z_{t})$.
                    \STATE Add transition to replay buffer.
                \ENDFOR
                \STATE Update parameters $\theta$ of $\pi_{\theta}$ with the policy gradient loss.
            \ENDWHILE
            \begin{center}
  \textit{**** testing phase begins ****}
            \end{center}
            \FOR{each input $z$ in $\mathcal{Z}_{\textrm{test}}$}
                \STATE Get state $s \leftarrow \mathrm{PLM}(z)$. 
                \STATE Get final prediction according to Eq. \ref{eq:final}.
            \ENDFOR
            \ENSURE A trained policy network $\pi_{\theta}$, predictions for test inputs.
	\end{algorithmic}
\end{algorithm}

\fakeparagraph{Model Overview} In response to these challenges, we define the discrete prompt matching problem as a Reinforcement Learning (RL) problem, \textit{Markov Decision Process (MDP)}, as shown in Algorithm \ref{alg:2}. 
For the action space $A$ of the RL agent, an action $a_k$ denotes that the agent selects a prompt $x_k$ from the prompt set $\mathcal{X}$ obtained in the  \textit{prompt set construction} stage.

At each step $t$ of the training phase, given a state $s_{t} = \mathrm{PLM}(z_{t})$, \ie the last hidden layer embedding of input $z_{t}$,  the RL agent takes an action $a_t$ of selecting a prompt $x_{t}$ from the prompt set $\mathcal{X}$ according to the policy $\pi_{\theta}(a_{t}|s_{t})$ where $\theta$ is the learnable parameter of the policy network.
We concatenate $x_{t}$ with $z_{t}$ and input them into PLM to complete downstream tasks, and calculate the reward $r_{t}$ of the RL agent based on the output of the PLM.
The goal of the RL agent is to maximize the expected reward $R=\mathbb{E}(\sum_{t=1}^{T}\gamma ^{t}r_{t})$, where $\gamma ^{t}$ is the discount factor at step $t$.

In the testing phase, we adopt the \textit{ensemble decision-making approach} for prompt selection. 
The prompts with  top-$k$ probability values are then entered into PLM to perform downstream tasks, which are weighted by the probability from the policy network $\pi_{\theta}$. Given an input $z$ and its corresponding state $s$, the final prediction obtained by \sysname{} at label $c$ can be expressed as
\begin{align}
    P(c|z) = \mathrm{softmax}(\sum_{j = 1}^{k}\pi_{\theta}(a_{j}|s)\log(p_{\mathrm{LM}}(c|x_{j},z))) \label{eq:final}
\end{align} 

\fakeparagraph{State Space}
In reinforcement learning, the concept of state space describes all the information about an environment at a given point in time. 
PLM is pre-trained on a large-scale unlabeled corpus based on self-supervised learning, allowing the model to capture complex language patterns, including long-distance dependence, polysemy disambiguation, sentence structure, etc \cite{dong2019unified, clark2020electra}.
In this work, we use the last hidden layer embedding of the outputs in the PLMs to represent the state $s$, which is subsequently input into the policy network.  
To ensure that the difference between states is distinguishable to the RL agent, we dynamically maintain a mean and standard deviation during training of the policy network to normalize observations of the state.

\fakeparagraph{Action Space}
An action $a \in A$ is proposed to match an appropriate prompt for an input based on the observed state, where the action space size $|A| = h$.
To make action decisions, we train a policy network  $\pi_{\theta}(.)$, which is a simple two-layer fully connected network, and parameters $\theta$ are optimized by the policy gradient algorithm \cite{sutton1999policy}. 
For input $z_{t}$, $\pi_{\theta}(.)$ outputs the  probability distribution of actions by
\begin{align}
    \pi_{\theta}(s_{t})=\mathrm{softmax}(w_{2}\cdot\tanh(w_{1}\cdot s_{t}))
\end{align}
where $w_{1}$ and $w_{2}$ represent the parameters $\theta$ of the two fully connected layers.

\fakeparagraph{Reward Design}
The reward received by the RL agent acts as the feedback that directly guides the update direction of the policy network. 
In this work, we aim to ensure that the RL-agent-selected prompts for the inputs can accurately complete the downstream task while maintaining balanced predicted label distribution.
To achieve this, we re-use the \metricname{} score to evaluate the degree of match between the prompt and input. 
Specifically, given an input $z_{t}$, we calculate \metricname{}$(x_{t}, z_{t})$ as the step reward $r_{t}$ of the RL agent after selecting the prompt $x_{t}$.

The reward scale obtained by RL agents can vary greatly due to disparities among different inputs. As a result, RL agents may overly focus on certain inputs during the training phase and become trapped in local optima. To address this issue, we normalize all rewards $r_{t}$ of the RL agent during training to maintain a relatively stable scale.

\begin{table*}[h]
\centering
\renewcommand\arraystretch{1.2}
\begin{threeparttable}[b]{
\begin{tabular}{ll|ccccp{1.25cm}<{\centering}}
\toprule
\textbf{Category}&\textbf{Method}               & \textit{\textbf{SST-2}}       & \textit{\textbf{MR}}          & \textit{\textbf{CR}}          & \textit{\textbf{Yelp}}       & \textit{\textbf{Avg.}} \\
\midrule 
\multirow{3}{4em}{Continuous Prompt} & Soft   Prompt Tuning & 73.84 $\pm$ 10.9 & 74.17 $\pm$ 14.6 & 75.89 $\pm$ 11.8 & 88.76 $\pm$ 4.73  & 78.17 \\
&Black-Box   Tuning   & 89.11 $\pm$ 0.92  & 86.60 $\pm$ 1.32  & 87.45 $\pm$ 1.06  & 93.22 $\pm$ 0.54 & 89.10 \\
&AutoPrompt           & 75.04 $\pm$ 7.64  & 62.02 $\pm$ 0.85  & 57.53 $\pm$ 5.88  & 79.81 $\pm$ 8.39 & 68.60 \\
\midrule
\multirow{6}{4em}{Discrete Prompt} &Manual   Prompt \tnote{$\dagger$}      & 82.82  $\pm$ 0.00      & 80.88 $\pm$ 0.00        & 79.60 $\pm$ 0.00        & 83.01   $\pm$ 0.00    & 81.58 \\
&Instruction \tnote{$\dagger$}            & 89.03 $\pm$ 0.00       & 85.18 $\pm$ 0.00        & 80.81 $\pm$ 0.00       & 84.44 $\pm$ 0.00      & 84.87 \\
&In-Context   Demo    & 85.91 $\pm$ 0.72  & 80.58 $\pm$ 1.44  & 85.50 $\pm$ 1.52  & 89.67 $\pm$ 0.48 & 85.42 \\
&GrIPS                & 87.14 $\pm$ 1.57  & 86.11 $\pm$ 0.33  & 80.02 $\pm$ 2.57  & 88.23 $\pm$ 0.17 & 85.38 \\
&RLPrompt \scriptsize{(SOTA)}     & 90.87 $\pm$ 0.86  & 86.85 $\pm$ 0.51  & 89.62 $\pm$ 1.36  & 93.78 $\pm$ 2.98 & 90.28 \\
\cmidrule(lr){2-7}
 &\sysname{}       & \textbf{93.62 $\pm$ 0.72}  & \textbf{88.58 $\pm$ 0.91}  & \textbf{90.76 $\pm$ 0.50}  & \textbf{94.25 $\pm$ 0.41} & \textbf{91.80} \\
\bottomrule
\end{tabular}
\caption{Comparison of the accuracy of \sysname{} and baseline methods on few-shot text classification tasks. The last column shows the average accuracy of each method on the four datasets. Overall, the \sysname{} method outperforms baseline methods in all cases. $\dagger$ Methods not affected by random seeds.}
\label{main_exp}
}
\end{threeparttable}
\end{table*}

\fakeparagraph{Other Key Details}
During the training process, we utilize the policy gradient algorithm to update the policy network. 
To enhance the algorithm's exploratory potential and accelerate the convergence speed, we follow  \citet{sutton1988learning} and incorporate entropy into the loss computation of the strategy network. 
This inclusion allows the policy network to continually optimize the primary loss while maximizing the entropy of the strategy, thereby minimizing the possibility of the strategy succumbing to local optimum solutions. 
Additionally, we use the constant decay method \cite{tesauro1991practical} to control the learning rate of the policy network, which helps the algorithm to converge faster in the early stage of training and optimize the model more stably in the later stage.


\section{Experiments}
To demonstrate the effectiveness of \sysname{}, we conduct extensive experiments on four open-source datasets of sentiment classification tasks, including \textbf{SST-2} \cite{socher2013recursive}, \textbf{Yelp} \cite{zhang2015character}, \textbf{MR} \cite{pang2005seeing}, and \textbf{CR} \cite{hu2004mining}, and three tasks of \textbf{GLUE} \cite{wang2018glue} in the few-shot setting.
We also analyze the superiority of \sysname{} from various aspects: 
a) \textbf{Ablation experiments} to analyze the effect of modules in \sysname{} on downstream tasks; 
b) \textbf{Universality} in few-shot settings;
c) \textbf{Robustness} to choice of verbalizers;
d) \textbf{Generalization} for PLMs with different sizes;
e) \textbf{Lightweight} and \textbf{Efficiency} method deployment.



 



\subsection{Experiment Settings}
The setting of comparison experiments, including competitors and our model \sysname{}, follows \citet{deng2022rlprompt}. 
Also, we utilize a few-shot experiment following  \citet{perez2021true}, \ie randomly select 16 samples from each category $c$ of the dataset as the training set.
Meanwhile, we use the same sampling method for the validation set. 
Therefore, the size of our training and validation sets is $16\times \left | \mathcal{C} \right |$. 

We chose RoBERT-large \cite{liu2019roberta} for all downstream tasks. 
And we use GPT-4 \cite{OpenAI2023GPT4TR} API to generate 60 prompts on each dataset, screening out 15 of them as action spaces for reinforcement learning. 
In the policy network, $w_{1}\in \mathbb{R}^{1024\times 600}$ and $w_{2}\in \mathbb{R}^{600\times 15}$. 
We use AdamW with eps of 0.00001 during training of 200 epochs. The learning rate is 0.001,  and mini-batch size is 32. More details are shown in the appendix.

\subsection{Competitors}
The baselines for comparison are as follows:

\noindent \textbf{Soft Prompt Tuning } \cite{lester2021power} replaces discrete prompt tokens with learnable feature vectors, and optimizes prompt through gradient information of PLMs.

\noindent \textbf{Black-Box Tuning} \cite{sun2022black} combines the characteristics of discrete and continuous prompt optimization methods, optimizing the sequence of continuous prompt tokens attached to PLMs inputs without gradient.

\noindent \textbf{AutoPrompt} \cite{shin2020autoprompt} performs multiple rounds of iteration based on gradient information, replaces the vocabulary in the prompt, and optimizes the discrete prompt template.

\noindent \textbf{Manual Prompt} applies the prompt designs of \citet{bach2022promptsource}, directly combines the prompt with the input for downstream tasks.

\noindent \textbf{Instruction} is a basic form of discrete prompting that facilitates PLMs to complete downstream tasks through an explanatory text. 
We design prompts for each task according to \citet{mishra2021cross}.

\noindent \textbf{In-Context Demo} \cite{brown2020language} randomly selects training data as examples to prompt PLMs to process subsequent input.

\noindent \textbf{GrIPS} \cite{prasad2022grips} optimizes discrete prompts by lexical-level editing on basic prompts, \ie{} substitution, deletion, and swapping, \etc

\noindent \textbf{RLPrompt} \cite{deng2022rlprompt} uses reinforcement learning techniques to individually train partial parameters of PLMs to generate discrete prompts for PLMs on downstream tasks.

\subsection{Performance Comparison}
As shown in \tabref{main_exp}, the \sysname{} method outperforms its competitors on all datasets. 
Specifically, compared with the SOTA method RLPrompt, \sysname{} achieves accuracy improvements of \textbf{2.75\%}, \textbf{1.73\%}, and \textbf{1.14\%} on SST-2, MR, and CR datasets, respectively. 
Additionally, on the Yelp dataset, \sysname{} still achieved a \textbf{0.47\%} performance improvement with greater stability, despite RLPropmt performing well.
Furthermore, compared with other prompt optimization methods using sorely supervision (\ie AutoPrompt and GrIPS), \metricname{}, which combines the unsupervised and supervised components excels. 
In terms of average accuracy over all datasets, \sysname{} performs \textbf{23.20\%} better than AutoPrompt and \textbf{6.42\%} better than GrIPS in accuracy.
Compared to Soft Prompt Tuning, one of the most popular prompt optimization methods, \sysname{} achieves \textbf{13.63\%} better accuracy on all four datasets while ensuring prompt readability. 
Moreover, our proposed multi-round dialogue alignment strategy can build the high-quality prompt set stably, resulting in a smaller standard deviation of \sysname{}'s performance compared to Soft Prompt Tuning.



\subsection{Ablation Study}
To study the impact of each component of \sysname{} on the final performance,  we perform ablation experiments on \textit{generation strategy}, \textit{selection metric}, and \textit{matching strategy}.

\fakeparagraph{Generation Strategy}
We compare the prompt generation strategy of \sysname{} with two commonly used strategies: \textit{Examples-Only} and \textit{Prompt-Examples} \cite{ubani2023zeroshotdataaug, min2022rethinking, dai2023chataug}.
\textit{Example-Only} prompt generation strategy first concatenates a certain number of inputs into a piece of text in random order, then enters the text into GPT-4 in a single round of dialogue for generating the pseudo-label inputs. 
\textit{Prompt-Examples} strategy is based on the \textit{Examples-Only} strategy, applying an explanatory text prefix to the input combination text. 
The prefix usually contains introductions and requirements for the downstream tasks. 
In the experiment, we use the same training data, utilize different generation strategies to generate 20 pseudo-label inputs as prompts and calculate their average accuracy on the test set. 
We provide the specific input used by the three prompt generation strategies in the appendix.

\begin{table}[h]
\centering
\begin{tabular}{lcccc}
\toprule
\textbf{Method}           & \textbf{\textit{SST-2}} & \textbf{\textit{MR}}   & \textbf{\textit{CR}} & \textbf{\textit{Yelp}}  \\
\midrule
Examples Only   & 86.34  & 75.80 & 80.16 & 90.60 \\
Prompt Examples & 78.07  & 83.23 & 88.43 & 87.95\\
\sysname{} Examples  & \textbf{89.46}  & \textbf{85.69} & \textbf{88.99} & \textbf{91.67} \\
\bottomrule
\end{tabular}
\caption{
Comparison of prompt generation strategies. \sysname{} Examples are generated via the \textit{multi-round dialogue alignment strategy}.
}
\label{abl_gen}
\end{table}

\tabref{abl_gen} demonstrates that the \sysname{}'s prompt generation strategy, \ie \textit{multi-round dialogue alignment strategy}, results in an average accuracy improvement of \textbf{3.12\%}, \textbf{2.46\%}, \textbf{0.56\%} and \textbf{1.07\%} on the SST-2, MR, CR and Yelp datasets than the best comparison strategy, respectively.
We also found that \textit{Examples-Only} and \textit{Prompt-Examples} strategies show significant performance fluctuations when the dataset changes. In contrast, our \textit{multi-round dialogue alignment} strategy is much more stable, indicating that \sysname{} generates a superior set of prompts by aligning with the training set data via GPT-4 dialogue.

\fakeparagraph{Selection Metric}
Our prompt screening metric \metricname{} consists of two parts: \textit{supervised information} and \textit{unsupervised information}. 
To evaluate the effectiveness of each component, we compare \metricname{} to use these two sole parts. 
We select the top-15 prompts with the highest scores on each metric from the same prompt set and then calculate their average accuracy on the test set. 

\begin{table}[h]
\centering
\begin{tabular}{lcccc}
\toprule
\textbf{Method}           & \textbf{\textit{SST-2}} & \textbf{\textit{MR}}   & \textbf{\textit{CR}}  & \textbf{\textit{Yelp}} \\
\midrule
Supervised   & 85.07          & 78.45          & 87.55   & 90.78       \\
Unsupervised & 87.13          & 78.37          & 87.35   & 91.32       \\
\metricname{} in \sysname{}      & \textbf{87.72} & \textbf{78.60} & \textbf{88.01} & \textbf{92.71} \\
\bottomrule
\end{tabular}
\caption{Ablation study on selection metrics.
} 
\label{abl_sel}
\end{table}

\tabref{abl_sel} demonstrates the superior performance of \metricname{} in prompt screening. 
For example, on the SST-2, the average accuracy of the prompts screened by SUE is \textbf{0.59\%} higher than that of the best-performing comparison metric.

It is noteworthy that prompts selected solely using \textit{unsupervised information} achieves comparable performance to \textit{ supervised information}. This finding indicates that \sysname{} can potentially perform well on zero-shot tasks. 

\fakeparagraph{Matching Strategy}
To prove the superiority of utilizing reinforcement learning in matching prompts, we compare it with the two other prompt matching methods, \ie \textit{Random} and \textit{Similarity-based}.
The \textit{Random} method randomly matches the prompt and the input, while the \textit{Similarity-based} method matches them based on the cosine similarity between the inputs and the prompt feature embeddings.

\begin{table}[h]
\centering
\begin{tabular}{lcccc}
\toprule
\textbf{Method}           & \textbf{\textit{SST-2}} & \textbf{\textit{MR}}   & \textbf{\textit{CR}} & \textbf{\textit{Yelp}}  \\
\midrule
Random             & 92.13          & 87.34          & 89.50   & 92.03       \\
Similarity-based   & 91.38          & 88.60          & 89.33    & 92.61      \\
RL in \sysname{} & \textbf{94.03} & \textbf{89.07} & \textbf{90.95} & \textbf{94.32} \\
\bottomrule
\end{tabular}
\caption{Comparison of the matching strategies.
}   
\label{abl_match}
\end{table}

As shown in \tabref{abl_match}, the matching method for RL in \sysname{} achieves the best performance, \eg a \textbf{1.90\%} improvement in accuracy on SST-2.
It indicates our RL agent can capture the implicit connection between the prompt and the input while matching.

\subsection{Discussions}

\fakeparagraph{Analysis on Universality}
To demonstrate \sysname{}'s universality in few-shot settings, we compared it with baseline methods on the GLUE \cite{wang2018glue} natural language inference and reading comprehension task, using the base template of \citet{gao2020making}.
Lacking settings and design of some aforementioned baseline methods on these tasks, here we compared with Soft Prompt Tuning, Black-Box Tuning, Manual Prompt, and In-Context Demo.

As shown in Table \ref{universality}, results show that \sysname{} outperforms all baseline methods significantly, including the prevailing methods, Soft Prompt Tunning \cite{lester2021power} and Black-Box Tuning \cite{sun2022black}. 
\eg{} \sysname{}, achieves a performance gain of \textbf{2.7\%} in the QNLI task, and the improvement reaches an astonishing \textbf{5.4\%} in the MRPC task. 
This results demonstrate that \sysname{}'s good universality in the few-shot setting across various tasks, which greatly stimulates the downstream ability of PLMs.

\begin{table}[]
\centering
\small
\renewcommand\arraystretch{1.2}
\begin{tabular}{lccc}
\toprule
\textbf{Method}           & \textbf{\textit{RTE}} & \textbf{\textit{QNLI}}   & \textbf{\textit{MRPC}}   \\
\midrule
Soft Prompt Tuning       & 54.7 $\pm$ 10.6      & 49.7 $\pm$ 1.73      & 51.6 $\pm$ 2.39      \\
Black-Box Tuning       & 52.9 $\pm$ 0.44      & 48.8 $\pm$ 0.61      & 61.6 $\pm$ 0.97      \\
Manual Prompt       & 51.6 $\pm$ 0.00      & 50.8 $\pm$ 0.00      & 61.1 $\pm$ 0.00      \\
In-Context Demo.       & 59.7 $\pm$ 0.85      & 52.4 $\pm$ 0.67      & 45.8 $\pm$ 0.80      \\
\sysname{} & \textbf{61.2 $\pm$ 0.81}      & \textbf{55.1 $\pm$ 0.39}      & \textbf{67.0 $\pm$ 1.03}      \\
\bottomrule
\end{tabular}
\caption{Analysis on model universality. We use the GLUE Benchmark\footnote{https://gluebenchmark.com/} online evaluation, whose results are three-digit decimal numbers.}
\label{universality}
\end{table}

\fakeparagraph{Analysis on Robustness}
Prompt-based methods must map the verbalizer probabilities from PLMs' output into the label space that downstream tasks require. 
Therefore, the choice of verbalizer directly affects the final performance of PLMs. 
Previous work has discussed choosing suitable verbalizers for PLMs. 
Here we focus on the robustness of \sysname{} when facing different verbalizer choices, as the results shown in \tabref{robust}. 
We follow the experimental setup of RLPrompt \cite{deng2022rlprompt}.
Experiments show that \sysname{} outperforms Manual Prompt at three different verbalizer settings significantly. 
Meanwhile, compared to the SOTA method RLPrompt, \sysname{} also surpasses it slightly, which accounts for \sysname{}'s better robustness to the choice of verbalizer.

\begin{table}[h]
\resizebox{\columnwidth}{!}{
\centering
\small
\renewcommand\arraystretch{1.2}
\begin{tabular}{lccc}
\toprule
\textbf{Verbalizer}         & \textbf{Manual}        & \textbf{RLPrompt}            & \textbf{\sysname{}}       \\
\midrule
\texttt{bad/good}            & 79.73          & 91.22 $\pm$ 1.46          & \textbf{91.96 $\pm$ 0.41}          \\
\texttt{negative/positive} & 76.89          & 92.20 $\pm$ 0.65          & \textbf{93.64  $\pm$ 0.77}          \\
\texttt{terrible/great}      & 82.86        & 92.81 $\pm$ 0.85            & \textbf{93.58 $\pm$ 0.51} \\
\bottomrule
\end{tabular}
}
\caption{Analysis on \sysname{}'s robustness to verbalizers.
}
\label{robust}
\end{table}

\vspace{-10pt}
\fakeparagraph{Analysis on Generalization}
We analyze the model generalization for PLMs with different sizes, which is involved in two modules of \sysname{}: \textit{prompt generalization} and \textit{policy generalization}. 

First, the prompts generated by \sysname{} can transfer between PLMs of different sizes. That is, the prompts computed \metricname{} and selected based on a smaller PLM can also achieve good performance for downstream tasks in another larger PLM.

\begin{table}[h]
\centering
\begin{tabular}{lcccc}
\toprule
\textbf{Method}           & \textbf{\textit{SST-2}} & \textbf{\textit{MR}}   & \textbf{\textit{CR}} & \textbf{\textit{Yelp}}   \\
\midrule
Manual Prompt       & 82.82      & 80.88      & 79.60  & 83.01    \\
\sysname{} \scriptsize{generalized}         & 83.33      & 80.38      & 84.66  & 89.26     \\
\sysname{}       & 93.62   & 88.58    & 90.76    & 94.25  \\
\bottomrule
\end{tabular}
\caption{Analysis on generalization ability of \sysname{}'s prompts on different size PLMs. }
\label{general1}
\end{table}

In this experiment, we use the metric scores output from the RoBERTa-base (110M parameters) to select the prompts and test their generalization on RoBERTa-large (354M parameters) to get the downstream task prediction accuracy.
Impressively, \tabref{general1} shows that the prompts selected by smaller PLMs are well-transferable with a minor decline in accuracy than the vanilla, still achieving comparable performance to the Manual Prompt baseline. 



The \textit{policy generalization} concerns whether the trained policy network of \sysname{} can function well on different PLMs. 
We train a policy network on RoBERTa-base and apply it to RoBERTa-large. 
In this test, we keep the prompts unchanged and only focus on evaluating the policy's performance. 
\tabref{general2} shows that even if using a smaller model to train the policy network, its performance on the large model version is still better than the commonly used random policy. Also, generalized \sysname{} only shows a slight decrease in accuracy to the vanilla.

\begin{table}[h]
\centering
\begin{tabular}{lcccc}
\toprule
\textbf{Method}           & \textbf{\textit{SST-2}} & \textbf{\textit{MR}}   & \textbf{\textit{CR}} & \textbf{\textit{Yelp}}  \\
\midrule
Random       & 88.48      & 85.07      & 86.00   & 90.22   \\
\sysname{} \scriptsize{generalized} & 89.34     & 86.40      & 87.12  & 91.36    \\
\sysname{}       & 93.62   & 88.58    & 90.76    & 94.25  \\
\bottomrule
\end{tabular}
\caption{Analysis on generalization ability of the policy. 
}
\label{general2}
\end{table}

\vspace{-10pt}
\fakeparagraph{Analysis on Lightweight and Efficiency}
\sysname{} only needs to train a two-layer fully connected network for its policy network. 
The number of parameters is 0.62M, which is only \textbf{0.73\%} of the whole policy network (distilGPT-2 with 82.0M parameters and an additional MLP with 3.15M parameters) used by  RLPrompt  in the experiment. 

Meanwhile, as shown in \tabref{time}, we compare the time consumption of \sysname{} and the SOTA method RLPrompt on the SST-2 dataset using a single NVIDIA GeForce RTX 3090 GPU. 
We find that the compact action space design in \sysname{} dramatically reduces the training time, which is only \textbf{10.86\%} of RLPrompt's, while \sysname{}'s accuracy exceeds RLPrompt as mentioned in \tabref{main_exp}.
\begin{table}[h]
\centering
\begin{tabular}{lcc}
\toprule
\textbf{Metric}   & \textbf{RLPrompt} &
\textbf{\sysname{}}  \\
\midrule
Time per Iterator & 1.09 s & 1.01 s\\
Training Time  &  218.63 min & 23.75 min \\

\bottomrule
\end{tabular}
\caption{Time consumption on SST-2 dataset.}
\label{time}
\end{table}
\vspace{-10pt}

\section{Conclusion}
In this paper, we propose \sysname{}, a novel discrete prompt optimization method. 
To efficiently and accurately select high-quality prompts, we design a prompt generation strategy through multi-round dialogue alignment on GPT-4 and propose an efficient prompt evaluation metric, \metricname{}. 
In addition, we design a reinforcement learning framework based on policy gradients to match suitable prompts for a single input. 
Our experimental results demonstrate that \sysname{} significantly improves the performance of PLMs in various downstream tasks while ensuring prompt readability and transferability.
In subsequent analysis experiments, we also verify \sysname{}'s good universality, robustness, generalization ability, lightweight and efficiency. 

\section{Acknowledgments}
This work is supported by National Key R\&D Program (2020YFB1406900), National Natural Science Foundation of China (62272371, 62103323, U21B2018, 62161160337, 62132011, 62376210, 62006181, U20B2049), Initiative Postdocs Supporting Program (BX20190275, BX20200270), China Postdoctoral Science Foundation (2019M663723, 2021M692565), Fundamental Research Funds for the Central Universities under grant (xhj032021013, xtr052023004, xtr022019002), and Shaanxi Province Key Industry Innovation Program (2021ZDLGY01-02).



\bibliography{aaai24}

\clearpage

\newpage
\appendix
\section{Appendix}
\subsection{Hyperparameter}
We provide the training details of \sysname{} here. We optimize policy network parameters using the policy gradient algorithm and provide all hyperparameters for reference, which is shown in \tabref{hyper}. Among them, Entropy COE represents the weight coefficient of entropy in loss calculation. We find that appropriate Entropy COE can help improve the performance of RL agents.
\begin{table}[!h]
\centering
\renewcommand\arraystretch{1.3}
\tabcolsep=1pt
\begin{tabular}{l|ccccccc} 
\toprule
\textbf{\small Hyperparameter} & \textit{\textbf{SST-2}} & \textit{\textbf{Yelp}} & \textit{\textbf{MR}} & \textit{\textbf{CR}} & \textit{\textbf{RTE}} & \textit{\textbf{QNLI}} & \textit{\textbf{MRPC}}\\
\midrule
Learning Rate           & 1e-3&   1e-3&  1e-3&  1e-3  &  1e-3 &  1e-3 &  1e-3                                                                     \\
Batch Size              & 32    & 32    & 32    & 32  & 32 & 32 & 32                                                                \\
Entropy COE             & 0.059                   & 0.065                  & 0.060                & 0.068 & 0.050 & 0.055 & 0.059             \\
Action Dim              & 15    & 15    & 15    & 15 & 15 & 15  & 15                                                                     \\
Hidden Dim              & 600   & 600   & 600   & 600  & 600  & 600  & 600                                                                 \\
State Dim               & 1024  & 1024  & 1024  & 1024 & 1024  & 1024  & 1024                                                                  \\
Top- K                   & 10                      & 15                     & 7                    & 3  & 15 & 15    & 5             \\
$\lambda_1$               & 10   & 10   & 10   & 10  & 10  & 10 & 10                                                                   \\
$\lambda_2$               & 7.00                       & 6.50                    & 6.75                 & 6.75  & 6.00  & 6.50  & 6.50          \\
\bottomrule
\end{tabular}
\caption{Hyperparameters of \sysname{} in the main experiments.}
\label{hyper}
\end{table}

\subsection{Comparison of Different Prompting Methods}
We compare the properties of different methods to solve downstream tasks using PLMs. As shown in \tabref{comparison}, our proposed method \sysname{}  is the only one with all properties including frozen PLMs, guided optimize, gradient-free, zero-shot, prompt readable and prompt generalizable. 

\begin{table}[!h]
\small
\centering
\renewcommand\arraystretch{1.3}
\tabcolsep=1pt
\begin{tabular}{l|cccccc} 
\toprule
\textbf{\small Methods} & \makecell[c]{Frozen 
 \\ PLMs} & \makecell[c]{Guided 
 \\ Optimize} & \makecell[c]{Auto- 
 \\ mated} & \makecell[c]{Read- 
 \\ able} & \makecell[c]{Generaliz- 
 \\ able} & \makecell[c]{Gradient 
 \\ Free}\\
\midrule
\textit{Fine-Tuning}              &\color{red}\XSolidBrush  & \color{green}\Checkmark   & \color{red}\XSolidBrush   & \color{red}\XSolidBrush    & \color{red}\XSolidBrush  & \color{red}\XSolidBrush                                                   \\
\textit{Manual Prompt}          & \color{green}\Checkmark &   \color{red}\XSolidBrush &  \color{green}\Checkmark&  \color{green}\Checkmark &  \color{green}\Checkmark    & \color{green}\Checkmark                                                              \\
\textit{Instructions}              &\color{green}\Checkmark    & \color{red}\XSolidBrush    & \color{green}\Checkmark  &\color{green}\Checkmark &  \color{green}\Checkmark   & \color{green}\Checkmark                                                           \\
\textit{In-Context Demo.}             & \color{green}\Checkmark                  & \color{red}\XSolidBrush                  & \color{red}\XSolidBrush                & \color{green}\Checkmark  &  \color{green}\Checkmark    & \color{green}\Checkmark \\
\textit{Soft Prompt Tuning}              & \color{green}\Checkmark    & \color{green}\Checkmark    & \color{red}\XSolidBrush    & \color{red}\XSolidBrush  & \color{red}\XSolidBrush          & \color{red}\XSolidBrush                                                        \\

\textit{AutoPrompt}               &\color{green}\Checkmark  & \color{green}\Checkmark  & \color{red}\XSolidBrush  & \color{red}\XSolidBrush    &  \color{green}\Checkmark    & \color{red}\XSolidBrush                                           \\
\textit{\makecell[l]{GrIPS}}
&\color{green}\Checkmark 
&\color{red}\XSolidBrush
& \color{green}\Checkmark 
&\color{green}\Checkmark
& \color{green}\Checkmark
& \color{green}\Checkmark
\\
\textit{\makecell[l]{Black-Box Tuning}}
&\color{green}\Checkmark 
& \color{green}\Checkmark 
&\color{green}\Checkmark
&\color{red}\XSolidBrush
&\color{red}\XSolidBrush
& \color{green}\Checkmark
\\
\textit{RLPrompt}               &\color{green}\Checkmark  &\color{green}\Checkmark   &\color{green}\Checkmark   & \color{red}\XSolidBrush  &  \color{green}\Checkmark  & \color{green}\Checkmark\\      
\toprule                                    \textit{\sysname{} (mine)}               & \color{green}\Checkmark  & \color{green}\Checkmark   & \color{green}\Checkmark   & \color{green}\Checkmark &  \color{green}\Checkmark  & \color{green}\Checkmark\\
\bottomrule
\end{tabular}
\caption{Comparison of \sysname{} with various methods of using PLMs for downstream tasks.}
\label{comparison}
\end{table}

\subsection{Datasets}
In \tabref{datasets}, we provide details of the dataset used in the main experiment in the few-shot setting, including the type and size of the training, validation, and test sets. Our experiments include sentiment classification, natural language inference, reading comprehension and other tasks.

\begin{table}[!h]
\centering
\renewcommand\arraystretch{1.3}
\tabcolsep=1pt
\begin{tabular}{l|cccc} 
\toprule
\textbf{\small Datasets} & Type &  Class & Train/Validation & Test \\
\midrule
\textit{\textbf{SST-2}}           &Sentiment (Movie) &   2&  32&  1.8k                                                                     \\
\textit{\textbf{Yelp}}              &Sentiment (Yelp)    & 2    & 32  &38k                                                               \\
\textit{\textbf{MR}}             & Sentiment (Movie)                   & 2                  & 32                & 2k       \\
\textit{\textbf{CR}}              & Sentiment (Product)    & 2    & 32    & 2k                                                                     \\
\textit{\textbf{RTE}}              & NLI   & 2   & 32   & 3k                                                              \\
\textit{\textbf{QNLI}}               & QA/NLI  & 2  & 32  & 5.4k                                                                 \\
\textit{\textbf{MRPC}}               & Paraphrase   & 2   & 32   & 1.7k                                                                     \\
\bottomrule
\end{tabular}
\caption{Datasets of \sysname{} in the main experiments.}
\label{datasets}
\end{table}

\subsection{Prompt-based Settings}
As shown in \tabref{prompt_setting}, we show the base prompt template and label words used by different dataset inputs. $\mathrm{\left \langle S \right \rangle }$: input sentence. 
Note that the input of RTE, QNLI, and MRPC contains two independent sentences $\mathrm{\left \langle S_{1} \right \rangle }$ and $\mathrm{\left \langle S_{2} \right \rangle }$.

\begin{table}[!h]
\centering
\renewcommand\arraystretch{1.3}
\tabcolsep=1pt
\begin{tabular}{l|cc} 
\toprule
\textbf{\small Datasets} & Base template & Label words  \\
\midrule
\textit{\textbf{SST-2}}           & Reviews:$\mathrm{\left \langle S \right \rangle }$ Sentiment:[MASK] & positive/negative                                                   \\
\textit{\textbf{Yelp}}              & Reviews:$\mathrm{\left \langle S \right \rangle }$ Sentiment:[MASK] & positive/negative                                                              \\
\textit{\textbf{MR}}             & Reviews:$\mathrm{\left \langle S \right \rangle }$ Sentiment:[MASK] & positive/negative               \\
\textit{\textbf{CR}}              & Reviews:$\mathrm{\left \langle S \right \rangle }$ Sentiment:[MASK] & positive/negative                           \\
\textit{\textbf{RTE}}             & $ \mathrm{\left \langle S_{1} \right \rangle }$. [MASK], I believe $ \mathrm{\left \langle S_{2} \right \rangle }$  & Clearly/Yet                                                \\
\textit{\textbf{QNLI}}              & $\mathrm{\left \langle S_{1} \right \rangle }$? [MASK]. Yes, $ \mathrm{\left \langle S_{2} \right \rangle }$    & Okay/Nonetheless                                                \\
\textit{\textbf{MRPC}}               & $ \mathrm{\left \langle S_{1} \right \rangle }$. [MASK]! $ \mathrm{\left \langle S_{2} \right \rangle }$  & Rather/Alas                                       \\
\bottomrule
\end{tabular}
\caption{Prompt-based Setting of \sysname{} on each dataset}
\label{prompt_setting}
\end{table}

\subsection{Prompt Set $\mathcal{X}$}
We present the prompt sets generated and selected across all experimental datasets in Tables 14 to 20.
Observations indicate that the prompts formulated via the $\textsc{DP}_2\textsc{O}$ method exhibit superior readability. 
In addition, we also find that the semantics of some high-quality prompts are similar for a single dataset.

\subsection{Prompts for Generation}
In \tabref{tab:prompt_gpt}, we present the prompts inputted to GPT-4 for multi-round dialogue alignment strategy on each dataset.
At initialization, we use GPT-4 imitation to generate pseudo-labeled input.
Then we use the prompt to continuously alignment the pseudo-label input according to the newly provided inputs. 
$\mathcal{Z}_{\textrm{seed}}$: the high-quality input set selected from the training set with \metricname{} as the metric.

\clearpage

\begin{table*}[h]
\small
\centering
\renewcommand\arraystretch{1.2}
\begin{tabular}{p{1.0\linewidth}l}
\toprule
The high-quality prompt set $\mathcal{X}$ for SST-2\\ 
\midrule
\ttfamily  1. Review: the narrative fails to connect and feels like a missed opportunity. Sentiment: negative.\\ 
\ttfamily  2. Review: the movie lacks substance and relies heavily on stereotypes. Sentiment: negative.\\ 
\ttfamily  3. Review: the film tries too hard to be edgy, ultimately falling flat. Sentiment: negative.\\ 
\ttfamily  4. Review: the film\'s trite plot and subpar acting make for a tedious viewing experience. Sentiment: negative.\\ 
\ttfamily  5. Review: its shallow character development and predictable plot makes for a dull watch. Sentiment: negative.\\ 
\ttfamily  6. Review: The representation of cultural aspects was egregiously inaccurate and disrespectful. Sentiment: negative. \\
\ttfamily  7. Review: a cinematic delight that wins the hearts of the audience. Sentiment: positive. \\
\ttfamily  8. Review: The depiction of mental illness was stereotypical, and frankly offensive. Sentiment: negative. \\
\ttfamily  9. Review: it lacks depth, making it feel hollow and disconnected. Sentiment: negative. \\
\ttfamily  10. Review: the film's lackluster pacing and clumsy storytelling overshadow its potential. Sentiment: negative. \\
\ttfamily  11. Review: poor screenplay and uninspiring performances lead to a forgettable experience. Sentiment: negative. \\
\ttfamily  12. Review: a film that tickles your funny bone with its razor-sharp humor. Sentiment: positive. \\
\ttfamily  13. Review: The characters\' decisions in the plot were overly ridiculous and somewhat degrading. Sentiment: negative. \\
\ttfamily  14. Review: a movie that combines humor, emotion, and action in the perfect blend. Sentiment: positive. \\
\ttfamily  15. Review: a thrilling joyride that keeps viewers glued to their seats. Sentiment: positive. \\
\bottomrule
\vspace{-1em}
\end{tabular}
\caption{The high-quality prompt set $\mathcal{X}$ for SST-2.}
\label{tab:sst-2}
\end{table*}

\begin{table*}[!t]
\small
\centering
\renewcommand\arraystretch{1.2}
\begin{tabular}{p{1.0\linewidth}l}
\toprule
The high-quality prompt set $\mathcal{X}$ for CR\\ 
\midrule
\ttfamily  1. Review: It would not maintain a stable Bluetooth connection. Sentiment: Negative.\\ 
\ttfamily  2. Review: It wouldn't properly sync with my devices. Sentiment: Negative.\\ 
\ttfamily  3. Review: The smartwatch's operating system is rather unstable. Sentiment: Negative.\\ 
\ttfamily  4. Review: The PC operating system tends to crash often. Sentiment: Negative.\\ 
\ttfamily  5. Review: The OS of this smartwatch isn\'t user-friendly. Sentiment: Negative.\\ 
\ttfamily  6. Review: It wouldn't stop crashing during use. Sentiment: Negative. \\
\ttfamily  7. Review: It consistently fails to disconnect calls, much to my annoyance. Sentiment: negative. \\
\ttfamily  8. Review: The operating system the machine uses seems to have a few problems. Sentiment: negative. \\
\ttfamily  9. Review: It's not user-friendly at all. Sentiment: negative \\
\ttfamily  10. Review: The tablet's operating system is quite slow. Sentiment: Negative. \\
\ttfamily  11. Review: I must admit, the software running the gadget has several glitches. Sentiment: negative. \\
\ttfamily  12. Review: The phone's OS is not as smooth as I expected. Sentiment: Negative. \\
\ttfamily  13. Review: The device fails to disconnect calls properly. Sentiment: negative. \\
\ttfamily  14. Review: I will say that the OS that the phone runs does have a few issues. Sentiment: negative \\
\ttfamily  15. Review: The device simply won't end calls when needed. Sentiment: negative. \\
\bottomrule
\vspace{-1em}
\end{tabular}
\caption{The high-quality prompt set $\mathcal{X}$ for CR.}
\label{tab:cr}
\end{table*}

\clearpage
\begin{table*}[!t]
\small
\centering
\begin{tabular}{p{1.0\linewidth}l}
\toprule
The high-quality prompt set $\mathcal{X}$ for Yelp\\ 
\toprule
\ttfamily  1. Review: Far from the Vegas ambiance I anticipated. Bland and unexciting, much like an uninspiring suburb in Texas. The gaming machines were badly arranged, hampering the overall visual aesthetic of the place. Sentiment: negative.\\ 
\ttfamily  2. Review: Unacceptable service! I had to wait for an exorbitant amount of time for a simple transaction. The staff, especially Emily, were entirely unprofessional, chewing gum in plain sight with no consideration for their customers. Sentiment: negative.\\ 
\ttfamily  3. Review: Worst customer service experience! I was waiting for what felt like an eternity at the payment desk. The woman serving, Olivia, was outright unprofessional, constantly chewing gum and ignoring the needs of the customers. Sentiment: negative.\\ 
\ttfamily  4. Review: Be extremely cautious when making jewelry purchases here. I experienced frustrating delays in responses to my emails and calls. Moreover, the sales associate replaced the ruby we had selected with a garnet without our knowledge (we realized only when the item was delivered 6 weeks later). When the bracelet arrived, it lacked a gemstone and the store refused to refund our money, instead offering to attach the missing gemstone. Upon independent appraisal, the bracelet was only worth 30\% of what we shelled out. Utterly disappointed. Sentiment: negative.\\ 
\ttfamily  5. Review: Incredible Indian food. Some of the best tandoori chicken I've had on the East Coast. Dropped in for lunch with a colleague, and later came back for takeaway. The restaurant is large, so waiting should never be an issue. Quick service, friendly staff, reasonable prices, and exceptional dishes. I'll definitely return. Sentiment: positive.\\ 
\ttfamily  6. Review: Absolutely wonderful Italian restaurant. Definitely among the best pasta dishes I've tasted in the Northeast. Dropped in for a dinner with a friend and returned for takeout later. The restaurant is quite large, so waiting for a table should never be a problem. Service was swift and pleasant. Fair prices and excellent meals. Can't wait to go back. Sentiment: positive. \\
\ttfamily  7. Review: Exercise utmost caution while buying jewelry at this store. Experienced significant delays in response to phone calls and emails. The sales representative even swapped the emerald we purchased for a peridot without our consent (only noticed after it arrived 5 weeks later). The necklace arrived missing a gem, and the store refused to issue a refund, proposing to only replace the missing gem. On independent appraisal, the necklace was worth a mere 30\% of what we paid. A thoroughly disappointing experience. Sentiment: negative. \\
\ttfamily  8. Review: Absolutely horrendous customer service! 30 minutes to get a book from the library desk! Seriously?! This is unacceptable! The librarian was so unprofessional, she showed no regard for patrons and was busy on her personal laptop the whole time. Her name was Emily. Sentiment: negative. \\
\ttfamily  9. Review: Absolutely top-tier food. This is some of the best Indian food I've experienced in the Southwest. I stopped by for a relaxed dinner with my partner during the weekend and came back a few days later for pickup. The restaurant's interior is quite roomy, so I highly doubt there would be much wait for a table. Service was efficient and congenial. Prices were acceptable and we were utterly delighted with our meals, looking forward to digging into the leftovers. I will certainly go back. Sentiment: positive. \\
\ttfamily  10. Review: Atrocious service! Had to wait an absurd amount of time at the customer service desk for a basic transaction. The staff, especially Mark, was downright rude and unprofessional, chewing gum without any respect for customers. Sentiment: negative. \\
\ttfamily  11. Review: Terrific barbecue joint. Some of the best brisket and ribs I've tried in town. I stopped by for a quick lunch and found myself back for more a few days later. It's quite roomy, so you won't have to worry about waiting. Service was prompt and cordial. Prices were fair, and the food was great. I'll be back for sure. Sentiment: positive. \\
\ttfamily  12. Review: Fantastic food. Hands down, some of the best Vietnamese cuisine I've had in the Southeast. I went in for a quick lunch with a colleague on a weekday and returned a few days later for takeout. The interior is rather spacious, so waiting for a table seems unlikely. The service was swift and friendly. Prices were just right and we thoroughly enjoyed our meals, anticipating the leftovers. I am definitely going back. Sentiment: positive. \\
\ttfamily  13. Review: Really remarkable food. Some of the finest French cuisine I've had in the Northwest. I popped in for a swift lunch with a friend one afternoon and revisited a few days later for delivery. The venue is spacious, I don\'t think there would ever be a wait for a table. Service was prompt and cordial. Prices were sensible and our meals left us quite content, eagerly awaiting the leftovers. I will definitely be returning. Sentiment: positive. \\
\ttfamily  14. Review: I've had a wonderful experience with this airline. Flights are consistently on time, the customer service is responsive, and the baggage handling is excellent. This will be my go-to airline for future travel. Sentiment: positive. \\
\ttfamily  15. Review:Review: Really satisfying coffee. One of the best espresso I've had in this part of town. Dropped by for a quick caffeine fix in the afternoon and revisited for a to-go cup later in the week. The coffee shop has ample space, so finding a seat shouldn't be an issue. Quick service, pleasant staff, and reasonably priced. Will visit again. Sentiment: positive. \\
\bottomrule
\vspace{-1em}
\end{tabular}
\caption{The high-quality prompt set $\mathcal{X}$ for Yelp.}
\label{tab:yelp}
\end{table*}

\clearpage
\begin{table*}[!t]
\small
\centering
\renewcommand\arraystretch{1.2}
\begin{tabular}{p{1.0\linewidth}l}
\toprule
The high-quality prompt set $\mathcal{X}$ for MR\\ 
\toprule
\ttfamily  1. Review: a humdrum tale about bravery and camaraderie. Sentiment: negative.\\ 
\ttfamily  2. Review: visually stunning yet bereft of a compelling storyline. Sentiment: negative.\\ 
\ttfamily  3. Review: a dreary anecdote about sacrifice and resilience. Sentiment: negative.\\ 
\ttfamily  4. Review: crackerjack entertainment -- nonstop romance, music, suspense, and action. Sentiment: positive.\\ 
\ttfamily  5. Review: a half-hearted venture into the world of sci-fi. Sentiment: negative\\ 
\ttfamily  6. Review: a dull account of personal growth and discipline. Sentiment: negative. \\
\ttfamily  7. Review: a wearisome chronicle of integrity and determination. Sentiment: negative. \\
\ttfamily  8. Review: an uninspiring discourse on truth and morality. Sentiment: negative. \\
\ttfamily  9. Review: a dazzling portrayal of love, tragedy, comedy, and suspense. Sentiment: positive. \\
\ttfamily  10. Review: a monotonous lesson on trust and loyalty. Sentiment: negative. \\
\ttfamily  11. Review: a cinematic triumph — mesmerizing performances, absorbing screenplay, and beautiful score. Sentiment: positive. \\
\ttfamily  12. Review: a dry, academic dissection of human nature. Sentiment: negative \\
\ttfamily  13. Review: a tedious lecture on the dangers of greed. Sentiment: negative \\
\ttfamily  14. Review: a monotonous tale of perseverance and team spirit. Sentiment: negative. \\
\ttfamily  15. Review: a tiring treatise on the costs of ambition. Sentiment: negative \\
\bottomrule
\vspace{-1em}
\end{tabular}
\caption{The high-quality prompt set $\mathcal{X}$ for MR.}
\label{tab:mr}
\end{table*}

\begin{table*}[!t]
\small
\centering
\renewcommand\arraystretch{1.2}
\begin{tabular}{p{1.0\linewidth}l}
\toprule
The high-quality prompt set $\mathcal{X}$ for QNLI\\ 
\midrule
\ttfamily  1. Where can the tropical birds be found? Nonetheless. Yes, Many bird species prefer temperate climates.\\ 
\ttfamily  2. Who did the farmers collaborate with? Nonetheless. Yes, In most instances, local cooperatives played a significant role.\\ 
\ttfamily  3. Did the castle remain the center of town affairs after the King's departure? Nonetheless. Yes, The modern town hall was built, which attracted most of the town's administrative activities.\\ 
\ttfamily  4. Is it possible to grow crops in all climates? Nonetheless. Yes, Certain types of crops require specific environmental conditions to thrive.\\ 
\ttfamily  5. Is it possible to farm fish in every type of water body? Nonetheless. Yes, Certain fish species require specific water conditions to survive and reproduce.\\ 
\ttfamily  6. Can wave energy be harvested in any part of the ocean? Nonetheless. Yes, Wave power depends on wave height, speed, wavelength, and water density.\\
\ttfamily  7. What was the primary reason for implementing daylight saving time? Nonetheless. Yes, The practice was implemented to save energy and make better use of daylight during the evenings \\
\ttfamily  8. Why was the Clean Air Act passed in 1963? Nonetheless. Yes, This law was enacted to control air pollution on a national level \\
\ttfamily  9. What is the significance of the Kyoto Protocol? Nonetheless. Yes, The Kyoto Protocol was an international treaty committing state parties to reduce greenhouse gas emissions. \\
\ttfamily  10. What role did children play in the Industrial Revolution? Nonetheless. Yes, Children often worked in factories or mines, where they performed dangerous tasks for low wages. \\
\ttfamily  11. Why was the Berlin Wall constructed? Nonetheless. Yes, The Wall was built to prevent East Germans from fleeing to the West.\\
\ttfamily  12. What did the abbot remain as a town built around the abbey? Nonetheless. Yes, The proximity of the Palace of Westminster did not extend to providing monks or abbots with high royal connections; in social origin the Benedictines of Westminster were as modest as most of the order. \\
\ttfamily  13. What did the abbot remain as a town built around the abbey? Nonetheless. Yes, The proximity of the Palace of Westminster did not extend to providing monks or abbots with high royal connections; in social origin the Benedictines of Westminster were as modest as most of the order. \\
\ttfamily  14. How do vaccines help in the prevention of diseases? Nonetheless. Yes, Vaccination has been an effective method in controlling and eliminating many infectious diseases. \\
\ttfamily  15. What did the abbot remain as a town built around the abbey? Nonetheless. Yes, The proximity of the Palace of Westminster did not extend to providing monks or abbots with high royal connections; in social origin the Benedictines of Westminster were as modest as most of the order. \\
\bottomrule
\vspace{-1em}
\end{tabular}
\caption{The high-quality prompt set $\mathcal{X}$ for QNLI.}
\label{tab:qnli}
\end{table*}

\begin{table*}[!t]
\small
\centering
\renewcommand\arraystretch{1.2}
\begin{tabular}{p{1.0\linewidth}l}
\toprule
The high-quality prompt set $\mathcal{X}$ for RTE\\ 
\midrule
\ttfamily  1. The rise of online shopping has resulted in many traditional brick-and-mortar stores closing down. This trend is accelerating as the convenience of online shopping continues to appeal to consumers. Clearly, I believe E-commerce is changing the retail landscape.\\ 
\ttfamily  2. As scientific advancements are made, the threat of AI taking over jobs has become a reality. The rapid growth of automation in various sectors is inevitable. Clearly, I believe AI is transforming the job market.\\ 
\ttfamily  3. While we are aware of the harmful effects of smoking, tobacco use is still prevalent. Despite health warnings, many individuals continue to smoke. Clearly, I believe Tobacco use remains a major health issue.\\ 
\ttfamily  4. Some types of information spread faster on social media than others. Clearly, I believe Social media plays a significant role in information dissemination.\\ 
\ttfamily  5. Despite increased awareness and understanding, mental health continues to be a pervasive issue. Many individuals worldwide are still suffering from various mental health disorders. Clearly, I believe Mental health remains a major concern.\\ 
\ttfamily  6. Although promoting the importance of a healthy lifestyle is common, obesity rates worldwide are still on the rise. This is happening in spite of the availability of resources for maintaining a healthy weight. Clearly, I believe The fight against obesity is complex. \\
\ttfamily  7. Although we promote the virtues of a balanced diet, fast food chains are seeing an increase in sales. The appeal of quick, cheap meals is hard to resist. Clearly, I believe Fast food consumption is on the rise. \\
\ttfamily  8. Despite all the advancements in medicine, cancer remains a leading cause of death globally. Treatments have improved, but a definitive cure is still elusive. Clearly, I believe Cancer is a major global health concern. \\
\ttfamily  9. While efforts have been made to combat climate change, the increasing global temperature is proof of its continual presence. The impact of our actions on the environment is evident. Clearly, I believe Addressing climate change is a complex task.\\
\ttfamily  10. With the advent of smart technology, our reliance on electronic devices has increased tremendously. Despite concerns about digital dependency, device usage is increasing. Clearly, I believe We're becoming increasingly reliant on technology.\\
\ttfamily  11. As efforts to combat the spread of misinformation grow, the proliferation of fake news continues to be a problem. Social media platforms are struggling to filter out false information. Clearly, I believe Fake news is a persistent issue. \\
\ttfamily  12. Although modern society prides itself on progress, poverty is still a widespread issue. Inequality persists despite economic growth. Clearly, I believe Poverty is a persistent global issue. \\
\ttfamily  13. Some jobs are more prone to automation than others due to technological advancements. Clearly, I believe Job automation varies across different professions. \\
\ttfamily  14. Despite the known benefits of renewable energy sources, fossil fuels continue to dominate the energy market. This continues even as the effects of climate change become more apparent. Clearly, I believe Fossil fuels are still the primary energy source. \\
\ttfamily  15. Despite numerous safety measures, cyber attacks are becoming more sophisticated and frequent. The digital world is continually under threat. Clearly, I believe Cyber threats are becoming increasingly complex. \\
\bottomrule
\vspace{-1em}
\end{tabular}
\caption{The high-quality prompt set $\mathcal{X}$ for RTE.}
\label{tab:rte}
\end{table*}

\begin{table*}[!t]
\small
\centering
\renewcommand\arraystretch{1.2}
\begin{tabular}{p{1.0\linewidth}l}
\toprule
The high-quality prompt set $\mathcal{X}$ for MRPC\\ 
\midrule
\ttfamily  1. The profit margins for the next quarter are expected to be 20\% to 22\%. Alsa! That exceeds the company's initial projection of 18\% to 20\% profit margins.\\ 
\ttfamily  2. The revenue growth for this year is forecasted to be 5\% to 6\%. Alsa! That is higher than the company\'s previous forecast of 3\% to 4\% growth.\\ 
\ttfamily  3. Interest rates are expected to remain between 2.5\% and 2.7\% for the coming year.  Alsa! That is consistent with the central bank\'s earlier prediction of 2.5\% to 2.7\% interest rates.\\ 
\ttfamily  4. Despite rising by 2\% in early trading, BlueTech stocks fell to a weekly low.  Rather! BlueTech stocks experienced a 2\% rise before falling to a weekly low.\\ 
\ttfamily  5. The film opened with poor reviews but eventually garnered a large fan base.  Rather! Receiving poor reviews initially, the film later found great success with audiences.\\ 
\ttfamily  6. Smith\'s performance declined in the last quarter, recording a loss of 3\%. Rather! Recording a loss of 3\%, Smith\'s performance was down last quarter.\\
\ttfamily  7. The national debt grew by 7\% last quarter, hitting a record high. Rather! Last quarter, the national debt increased by 7\%, setting a new record. \\
\ttfamily  8. At 10:00 AM, the gold price was up \$5 at \$1,300, having previously reached \$1,305. Alsa! Gold prices rose \$5 to reach \$1,300, after previously hitting a high of \$1,305. \\
\ttfamily  9. Johnson criticized the policy, which he referred to as "a mistake", for causing economic decline. Rather! Referring to it as "a mistake", Johnson criticized the policy for leading to economic decline. \\
\ttfamily  10. The unemployment rate fell to 5.4\%, marking a three-year low. Rather! Falling to 5.4\%, the unemployment rate marked a three-year low. \\
\ttfamily  11. Around 09:00 PM, Xero stocks were down 5 points, or 2\%, at \$250, having earlier touched \$255. Alsa! Xero stocks dropped 5 points, or 2\%, to close at \$250, after touching \$255 earlier.\\
\ttfamily  12. The artist, known as "The Painter", unveiled a new series that challenged traditional forms. Rather! Known as "The Painter", the artist introduced a series that broke with tradition. \\
\ttfamily  13. At noon, EnergyX shares were up \$2, or 1.5\%, at \$150, after reaching a peak of \$151. Alsa! EnergyX shares climbed \$2, or 1.5\%, to set a record at \$150, after peaking at \$151. \\
\ttfamily  14. Harper, whom they call "The Analyst", provided a bleak forecast for the next quarter. Rather! Referred to as "The Analyst", Harper gave a pessimistic prediction for the next quarter. \\
\ttfamily  15. By 3:00 PM, AgriCorp\'s stocks were down 3\%, at \$30, having earlier fallen to \$29. Alsa! AgriCorp\'s stocks declined 3\%, closing at \$30, after earlier touching \$29. \\
\bottomrule
\vspace{-1em}
\end{tabular}
\caption{The high-quality prompt set $\mathcal{X}$ for MRPC.}
\label{tab:mrpc}
\end{table*}

\begin{table*}[!t]
\small
\centering
\begin{tabular}{p{1.0\linewidth}l}
\toprule
\textbf{SST-2}\\ 
\bottomrule
\ttfamily  Initialization: As a movie enthusiast, please generate 20 similar samples as shown in the parentheses.($\mathcal{Z}_{\textrm{seed}}[0:1]$)\\ 
\ttfamily  Alignment: Now imitate the example in parentheses, randomly changing the three samples generated by the previous dialogue, and the other samples remain unchanged.($\mathcal{Z}_{\textrm{seed}}[num]$)\\ 
\toprule
\textbf{Yelp}\\ 
\bottomrule
\ttfamily  Initialization: As a critic, please generate 20 similar samples as shown in the parentheses.($\mathcal{Z}_{\textrm{seed}}[0:1]$)\\ 
\ttfamily  Alignment: Now imitate the example in parentheses, randomly changing the three samples generated by the previous dialogue, and the other samples remain unchanged. ($\mathcal{Z}_{\textrm{seed}}[num]$)\\ 
\toprule
\textbf{MR}\\ 
\bottomrule
\ttfamily  Initialization: As a movie enthusiast, please generate 20 similar samples as shown in the parentheses.($\mathcal{Z}_{\textrm{seed}}[0:1]$)\\ 
\ttfamily  Alignment: Now imitate the example in parentheses, randomly changing the three samples generated by the previous dialogue, and the other samples remain unchanged. ($\mathcal{Z}_{\textrm{seed}}[num]$)\\ 
\toprule
\textbf{CR}\\ 
\bottomrule
\ttfamily  Initialization: As a customer, please generate 20 similar samples as shown in the parentheses.($\mathcal{Z}_{\textrm{seed}}[0:1]$) \\
\ttfamily  Alignment: Now imitate the example in parentheses, randomly changing the three samples generated by the previous dialogue, and the other samples remain unchanged.($\mathcal{Z}_{\textrm{seed}}[num]$) \\
\toprule
\textbf{RTE}\\ 
\bottomrule
\ttfamily  Initialization: As a prompt engineer, please generate 20 similar samples as shown in the parentheses. The form of prompt is Sentence1+Answer+Sentence2.($\mathcal{Z}_{\textrm{seed}}[0:1]$)\\
\ttfamily  Alignment: Now imitate the example in parentheses, randomly changing the three samples generated by the previous dialogue, and the other samples remain unchanged.($\mathcal{Z}_{\textrm{seed}}[num]$) \\
\toprule
\textbf{QNLI}\\ 
\bottomrule
\ttfamily  Initialization: As a prompt engineer, please generate 20 similar samples as shown in the parentheses. The form of prompt is Question+Answer+Sentence.($\mathcal{Z}_{\textrm{seed}}[0:1]$)\\
\ttfamily  Alignment: Now imitate the example in parentheses, randomly changing the three samples generated by the previous dialogue, and the other samples remain unchanged.($\mathcal{Z}_{\textrm{seed}}[num]$) \\
\toprule
\textbf{MRPC}\\ 
\bottomrule
\ttfamily  Initialization: As a prompt engineer, please generate 20 similar samples as shown in the parentheses. The form of prompt is Sentence1+Answer+Sentence2 and the answer there are only two answers: 'Alsa' or 'Rather'.($\mathcal{Z}_{\textrm{seed}}[0:1]$) \\
\ttfamily  Alignment: Now imitate the example in parentheses, randomly changing the three samples generated by the previous dialogue, and the other samples remain unchanged.($\mathcal{Z}_{\textrm{seed}}[num]$) \\
\bottomrule
\vspace{-1em}
\end{tabular}
\caption{Prompts for multi-round dialogue alignment strategy.}
\label{tab:prompt_gpt}
\end{table*}
\end{document}